\documentclass[letterpaper]{article} % DO NOT CHANGE THIS
\usepackage{aaai2026}  % DO NOT CHANGE THIS
\usepackage{times}  % DO NOT CHANGE THIS
\usepackage{helvet}  % DO NOT CHANGE THIS
\usepackage{courier}  % DO NOT CHANGE THIS
\usepackage[hyphens]{url}  % DO NOT CHANGE THIS
\usepackage{graphicx} % DO NOT CHANGE THIS
\usepackage{caption} % DO NOT CHANGE THIS AND DO NOT ADD ANY OPTIONS TO IT
\usepackage{algorithm}
\usepackage{algorithmic}

\usepackage{newfloat}
\usepackage{listings}
\DeclareCaptionStyle{ruled}{labelfont=normalfont,labelsep=colon,strut=off} % DO NOT CHANGE THIS
\floatstyle{ruled}
\newfloat{listing}{tb}{lst}{}
\floatname{listing}{Listing}

\usepackage{amsmath}
\usepackage{amsthm}
\usepackage{booktabs}
\usepackage{enumitem}
\usepackage{graphicx}
\usepackage{color}
\usepackage{algorithm} 
\usepackage{algorithmic} 
\usepackage{tabularx}
\usepackage{makecell}
\usepackage{multirow}
\usepackage{ragged2e}
\usepackage{bm}
\usepackage{url}
\usepackage{siunitx}  
\usepackage{booktabs}   
\usepackage{caption}   
\usepackage{array}     
\usepackage{multirow}   
\usepackage{graphicx} 
\usepackage{lipsum}
\usepackage[utf8]{inputenc} 
\usepackage[T1]{fontenc}  
\usepackage{url}         
\usepackage{booktabs}     
\usepackage{amsfonts}      
\usepackage{nicefrac}   
\usepackage{microtype}   
\usepackage{booktabs}
\usepackage{array}
\usepackage{siunitx}
\usepackage[table]{xcolor}  
\definecolor{tealgreen}{RGB}{0,100,100}
\usepackage[numbers,sort&compress]{natbib}
\usepackage[colorlinks=true,
            citecolor=tealgreen,  
            linkcolor=tealgreen, 
            urlcolor=tealgreen]   
            {hyperref}

\title{Whole-Body Coordination for Dynamic Object Grasping with Legged Manipulators}

\author{
    Qiwei Liang\textsuperscript{\rm 1,2},
    Boyang Cai\textsuperscript{\rm 2},
    Rongyi He\textsuperscript{\rm 2},
    Hui Li\textsuperscript{\rm 2},\\
    Tao Teng\textsuperscript{\rm 3},
    Haihan Duan\textsuperscript{\rm 4},
    Changxin Huang\textsuperscript{\rm 2},
    Runhao Zeng\textsuperscript{\rm 4}\thanks{Corresponding author: zengrh@smbu.edu.cn}
}

\affiliations{
    \textsuperscript{\rm 1}Hong Kong University of Science and Technology (Guangzhou)\\
    \textsuperscript{\rm 2}Shenzhen University
    \textsuperscript{\rm 3}T-Stone Robotics Institute, The Chinese University of Hong Kong\\
    \textsuperscript{\rm 4}Shenzhen MSU-BIT University\\
    \href{https://kolakivy.github.io/DQ/}{\textcolor{tealgreen}{https://kolakivy.github.io/DQ/}}
}

\begin{document}

\maketitle

\begin{abstract}
Quadrupedal robots with manipulators offer strong mobility and adaptability for grasping in unstructured, dynamic environments through coordinated whole-body control. However, existing research has predominantly focused on static-object grasping, neglecting the challenges posed by dynamic targets and thus limiting applicability in dynamic scenarios such as logistics sorting and human–robot collaboration. To address this, we introduce DQ-Bench, a new benchmark that systematically evaluates dynamic grasping across varying object motions, velocities, heights, object types, and terrain complexities, along with comprehensive evaluation metrics. Building upon this benchmark, we propose DQ-Net, a compact teacher–student framework designed to infer grasp configurations from limited perceptual cues. During training, the teacher network leverages privileged information to holistically model both the static geometric properties and dynamic motion characteristics of the target, and integrates a grasp fusion module to deliver robust guidance for motion planning. Concurrently, we design a lightweight student network that performs dual-viewpoint temporal modeling using only the target mask, depth map, and proprioceptive state, enabling closed-loop action outputs without reliance on privileged data. Extensive experiments on DQ-Bench demonstrate that DQ-Net achieves robust dynamic objects grasping across multiple task settings, substantially outperforming baseline methods in both success rate and responsiveness. 

\end{abstract}

\begin{figure*}[!t]
    \centering
    \includegraphics[width=\textwidth]{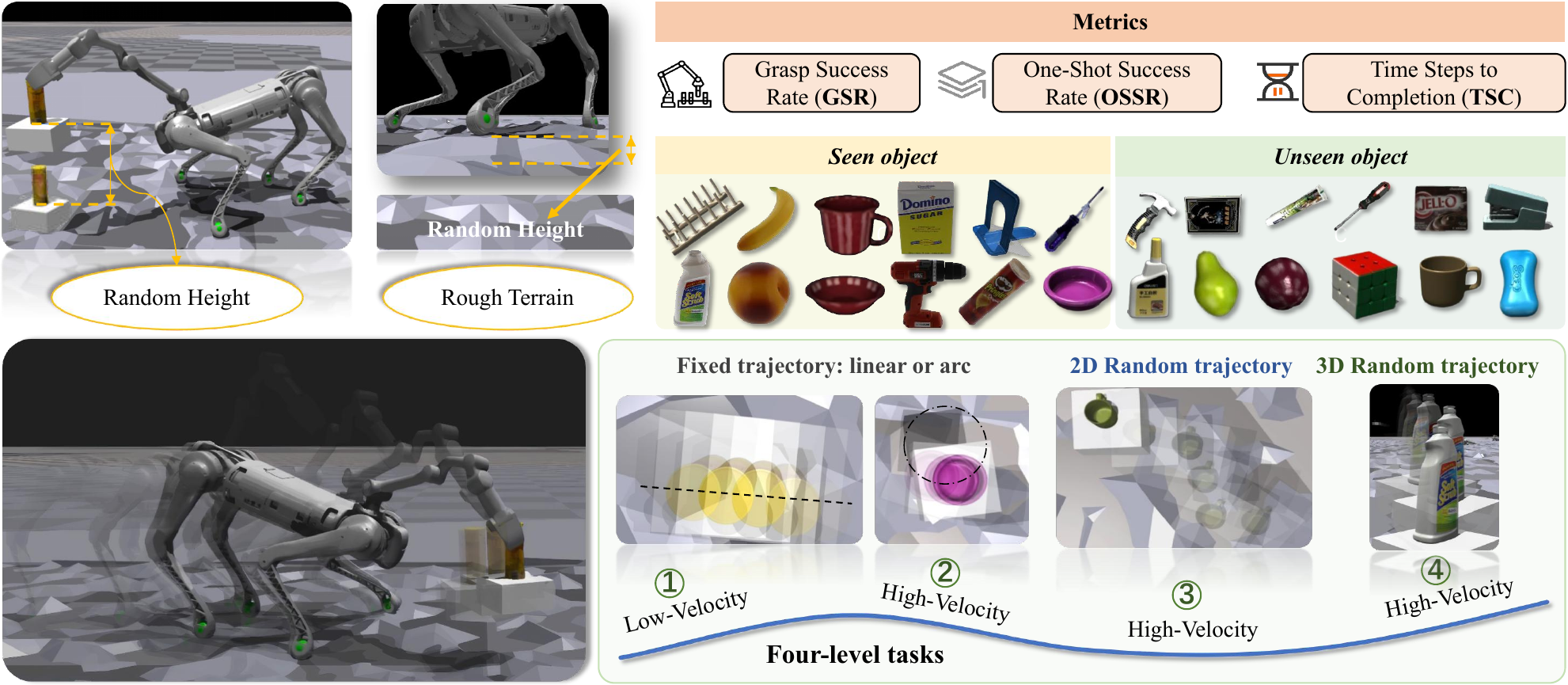}
    \caption{Overview of DQ-Bench: Our proposed benchmark provides a standardized and reproducible platform for evaluating dynamic object grasping with whole-body quadruped control. It systematically incorporates diverse grasping objects, multi-level task designs, and rigorous evaluation metrics to assess the adaptability, generalization, and efficiency of grasping strategies in dynamic scenarios.}
    \label{fig1:benchmark}
\end{figure*}

\section{Introduction}

Quadruped robots have emerged as a promising platform for mobile manipulation due to their superior mobility and terrain adaptability \cite{yang2022learning,youm2023imitating,Him2024,sun2024leg,mei2024quadrupedgpt,zeng2024video2reward}. With the integration of robotic arms, these systems are capable of performing manipulation tasks in complex environments \cite{bharadhwaj2024track2act,yokoyama2023asc,qiu2024wildlma,sleiman2021unified,ma2022combining,mittal2022articulated}, demonstrating significant potential in applications such as search and rescue, logistics, and human-robot collaboration. By coordinating multi-joint movements of the limbs and arm, quadrupeds can realize whole-body control, thereby enhancing both their dynamic responsiveness and operational workspace \cite{pan2024roboduet,portela2024whole,fu2022deep}.

While prior research has made substantial progress in manipulation with quadruped systems, most efforts focus on grasping static objects \cite{liu2024vbc,wang2024quadwbg,zhang2024gamma,qiu2024learning,ha2024umilegs}, assuming the target remains stationary. This simplification overlooks the critical challenges posed by dynamic object manipulation, where objects are in continuous motion—common in real-world scenarios such as conveyor-based logistics, interactive mobile targets, and human handovers \cite{xie2024gaprl,fang2023anygrasp}. These dynamic settings demand rapid perception and agile control, making stable and efficient grasping under motion- one of the core challenges in quadruped robot research \cite{huang2023earl,akinola2021dynamic}.

To study this problem, we introduce the first benchmark tailored for whole-body \textbf{d}ynamic grasping with \textbf{q}uadruped robots: \textbf{DQ-Bench}. This benchmark provides a reproducible and comprehensive evaluation framework to assess algorithmic generalization and decision efficiency under dynamic and challenging conditions. DQ-Bench supports physically realistic target motion, non-planar terrain, and multi-level task difficulties. It includes diverse common daily objects partitioned into seen and unseen sets to rigorously test perception robustness and policy transferability. Evaluation metrics include Grasp Success Rate (GSR), One-Shot Success Rate (OSSR), and Timesteps to Completion (TSC), jointly measuring grasping effectiveness, decision quality, and responsiveness.

Upon the benchmark, we seek to find a solution to the core problem of dynamic grasping. Unlike static manipulation, the key challenge here lies in maintaining whole-body stability while precisely controlling the arm’s end-effector in the presence of continuously changing relative motion. Even minor grasping pose deviations may lead to failure due to relative velocity between the robot and the target. Thus, accurately and efficiently predicting grasp poses at each time step from input signals is crucial yet highly challenging.
A straightforward approach is to employ high-performance grasping networks to decode object geometry and predict optimal grasp poses \cite{ma2023towards,qin2023rgb,chen2023keypoint,cai2022volumetric,dai2022dreds,wen2022transgrasp}. However, this is problematic in dynamic tasks: (i) these networks often rely on high-quality visual input, while quadrupeds typically observe targets from long distances with limited resolution; (ii) invoking such networks at every timestep incurs high computational costs, hindering real-time performance and training efficiency; (iii) conventional grasping networks output static grasp poses, lacking temporal consistency needed for continuous motion planning.

To overcome these limitations, we propose a framework for \textbf{d}ynamic object grasping with \textbf{q}uadruped robots, called \textbf{DQ-Net}. At the core of this framework is the \textbf{Grasp Fusion Module (GFM)}, designed to improve grasp quality and temporal consistency under motion. The GFM maintains a memory bank of multi-reference grasp poses, pre-generated in simulation based on object geometry and pose, and transforms these into a unified local coordinate system. At each timestep, the module constructs a query by using the object’s 6D pose and encoded point cloud, then the query will be matched against the memory bank via attention mechanisms to produce a refined and robust grasp pose for control.
This fused grasp pose, together with the object's motion state and robot proprioception, is fed into the policy head and trained via reinforcement learning to enable whole-body dynamic grasping. However, this pipeline relies on privileged information (object pose and velocity), which is challenging to obtain in practice due to the expense and measurement uncertainties of specialized sensors.

To bridge this gap, we further design a lightweight student policy network trained via knowledge distillation. It leverages only onboard sensory input, including dual-perspective visual observations (from the base and end-effector) and proprioceptive data. These inputs are separately encoded and fused using a dual-stream Transformer to capture both global semantic information and fine-grained object dynamics. The fused high-level features are decoded by the policy head to enable closed-loop dynamic grasping control. Extensive experiments on DQ-Bench demonstrate that DQ-Net significantly outperforms existing baselines across multiple dynamic tasks, achieving superior grasp success rates and faster response times. \textbf{Our contributions are summarized as follows:}
\begin{itemize}
    \item We construct \textbf{DQ-Bench}, the first benchmark for dynamic object grasping with quadruped robots, supporting realistic dynamics, diverse objects, multi-level task difficulty, and comprehensive evaluation across perception and control.
    \item We propose \textbf{DQ-Net}, a framework combining a grasp memory-based fusion module and a lightweight student network relying solely on dual-perspective vision and proprioception for stable and efficient whole-body dynamic grasping.
    \item We conduct extensive evaluations across challenging dynamic tasks, where DQ-Net consistently outperforms prior methods in terms of grasp success and policy responsiveness.
\end{itemize}

\section{Related Work}
\subsection{Whole-Body Control in Quadrupedal Manipulation}

Recent work has advanced unified whole-body control for quadrupedal robots integrating manipulation \cite{portela2024learning}. Fu et al.\cite{fu2022deep} propose Deep Whole-Body Control, a reinforcement learning policy coordinating both arm and leg joints for agile behaviors like picking and button-pressing. Liu et al.\cite{liu2024vbc} extend this with a vision-guided hierarchical policy that maps RGB-D inputs to whole-body trajectories, showing success on real hardware. However, these methods assume static targets and execute single-shot grasps without considering object motion. More recent approaches like QuadWBG~\cite{wang2024quadwbg} and GAMMA~\cite{zhang2024gamma} integrate grasp-aware planning into quadruped systems but still assume fixed objects during inference. Overall, existing whole-body control strategies excel under static assumptions but lack support for continuous target motion and dynamic tracking.

\subsection{Benchmarks for Robotic Grasping}

Large-scale datasets like GraspNet-1Billion~\cite{fang2020graspnet} and TARGO~\cite{xia2024targo} have enabled robust learning and evaluation of grasping models, yet both focus on fixed-base arms interacting with static objects. Efforts like DGBench~\cite{burgess2022dgbench} and GAP-RL~\cite{xie2024gaprl} target dynamic grasping by introducing moving objects and reactive grasp policies, but are still limited to tabletop settings with fixed manipulators and constrained motion. General benchmarks like ManiSkill~\cite{driess2021maniskill} and BEHAVIOR~\cite{jiang2025behavior} emphasize task diversity and mobile manipulation, yet lack support for dynamic grasping and do not involve locomotion-manipulation coupling. In contrast, our DQ-Bench introduces a physically realistic evaluation suite that combines non-planar terrain, 6-DoF target motion, and whole-body quadruped grasping, filling a critical gap in existing benchmarks.

\section{DQ-Bench: Dynamic Grasping Benchmark for Quadruped Robots}
Despite significant progress in static grasping tasks with quadruped robots, real-world applications often involve continuously moving targets that require real-time perception and grasping decisions. Currently, there is a lack of a standardized evaluation platform to systematically assess quadruped performance in dynamic grasping scenarios. To address this gap, we propose the first benchmark—\textbf{DQ-Bench}—for dynamic object grasping with full-body quadruped control. This benchmark provides a comprehensive and reproducible platform to evaluate the generalization and decision-making efficiency of various algorithms in complex dynamic environments. Our design systematically considers key factors from four dimensions—environment modeling, object selection, task difficulty levels, and evaluation metrics—ensuring high-fidelity modeling and rigorous assessment of dynamic grasping tasks. The overview of DQ-Bench is shown in Figure \ref{fig1:benchmark}.

\subsection{Environment Setup}

To ensure physical realism and support parallel training, we build the evaluation environment on the physics-based simulation platform \textit{Isaac Gym} \cite{makoviychuk2021isaac}. Objects are placed on a movable floating platform that follows predefined or randomized trajectories in 3D space. Compared to directly assigning random motion to objects, this method better reflects real-world object dynamics under external forces, avoiding unnatural motion that could distort grasping strategies. Additionally, we introduce uneven terrain to test the robot’s adaptability and whole-body coordination across varied ground conditions.

\subsection{Objects for Grasping}

We select representative objects from the YCB dataset \cite{YCB} that vary in shape, size, and weight to ensure task diversity and challenge. To test generalization, objects are split into two groups: \textit{seen} (used in both training and testing) and \textit{unseen} (used only during testing). This setup introduces real-world unpredictability and demands greater robustness and generalization from the grasping strategies.

\subsection{Multi-Level Task Design}

To comprehensively evaluate the quadruped's capability in dynamic grasping, we design four progressively challenging task levels based on object speed, trajectory complexity, terrain conditions, and spatial degrees of freedom:

\begin{itemize}
    \item \textbf{Level 1:} Object moves at low speed ($0\sim15$\,cm/s) along fixed trajectories (linear or arc).
    \item \textbf{Level 2:} Object moves at high speed ($15\sim30$\,cm/s) along fixed trajectories.
    \item \textbf{Level 3:} Object moves along random trajectories at high speed ($0\sim30$\,cm/s).
    \item \textbf{Level 4:} Object moves freely along the $z$-axis, resulting in random 3D trajectories.
\end{itemize}

These four levels span from 2D to 3D motion, flat to complex terrains, and low to high speeds. At each episode reset, the object’s initial height is randomly sampled within $[0.2, 0.7]$ meters, and the terrain is randomly varied within a flat range of $0$–$0.1$ meters.

\subsection{Evaluation Metrics}

To quantitatively assess grasping performance across task levels, we introduce three core metrics:

\begin{itemize}
    \item \textbf{Grasp Success Rate (GSR):} The proportion of successfully completed grasps, indicating the overall effectiveness of the strategy.
    \item \textbf{One-Shot Success Rate (OSSR):} The percentage of successful grasps completed in a single attempt without any re-adjustment, reflecting decisiveness.
    \item \textbf{Time Steps to Completion (TSC):} The number of time steps taken from task initiation to successful grasp, representing grasping efficiency.
\end{itemize}

These metrics jointly evaluate grasping strategies from the perspectives of success rate, decision quality, and response efficiency, providing insights into behavioral differences and performance bottlenecks in dynamic grasping scenarios.

\section{Proposed Method}

\begin{figure*}[!t]
    \centering
    \includegraphics[width=\textwidth]{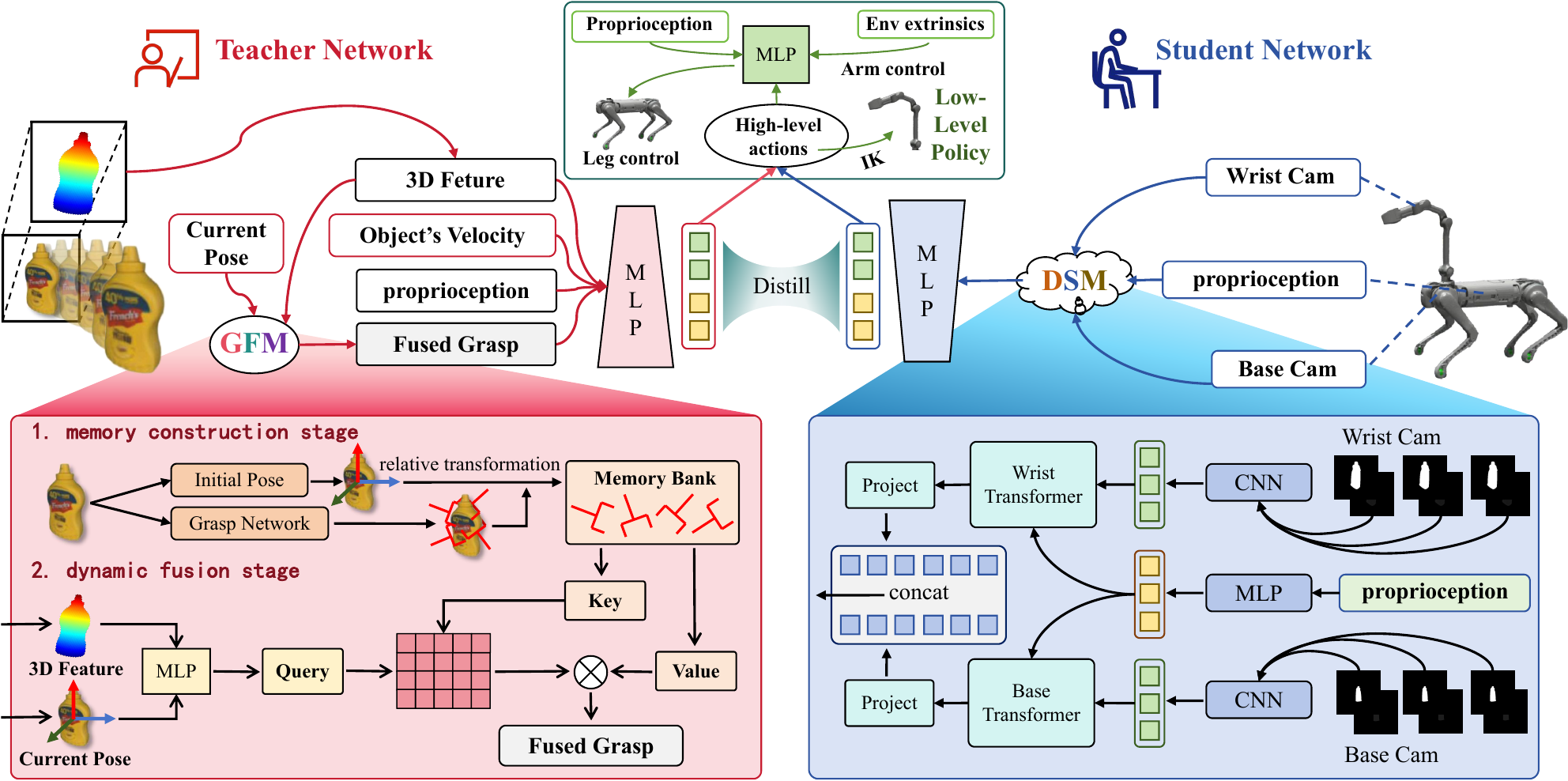}
    \caption{Overview of DQ-Net: We prepose a teacher-student framework for quadruped robots to perform dynamic object grasping. The teacher network takes point cloud features, object motion, robot proprioception, and a grasp representation from a Grasp Fusion Module (GFM), which builds a grasp memory and fuses object features with current pose via attention. The student network uses a dual-stream architecture to encode three-frame sequences of target masks and depth maps from wrist and base cameras. Features from both streams are fused with proprioception, processed by Transformers, and decoded into high-level actions. These high-level actions actions are further mapped to low-level control for end-to-end loco-manipulation.
    }
    \label{fig2:method}
    \vspace{-0.2cm}
\end{figure*}

\subsection{Overview}
We aim to develop a lightweight control system for quadruped robots that achieves whole-body coordination and dynamic grasping using only accessible perceptual inputs, such as depth maps and segmentation masks. The control framework comprises: \textbf{1)} a low-level policy (LLP) trained via reinforcement learning to produce full-body control signals from target velocity and end-effector poses; and \textbf{2)} a high-level policy (HLP) that maps perceptual observations to high-level commands for the LLP. 

However, training a vision-only policy via reinforcement learning is challenging due to the task’s complexity and the limited informativeness of raw visual inputs. To address this, we adopt a teacher-student distillation framework (Fig.~\ref{fig2:method}). The teacher is trained in simulation with privileged inputs (e.g., object pose and velocity) to learn a high-quality policy, while the student learns to imitate the teacher using only depth and segmentation maps.

While the teacher benefits from rich inputs, dynamic grasping still requires accurate and timely grasp pose estimation due to object movement. Traditional grasp pose estimation methods often incur high computational cost, suffer from latency, and rely on static visual inputs, which limits their effectiveness in long-range prediction and grasp pose refinement during approach.

In this paper, we introduce the Grasp Fusion Module (GFM) that dynamically selects grasp poses in real time. GFM maintains a memory bank of multi-directional grasp candidates and predicts the optimal grasp pose $\mathbf{g}_t$ using attention mechanisms, conditioned on point cloud features $\mathbf{f}_p$ and object pose $\mathbf{x}_t$. Then, $\mathbf{g}_t$ is fed into the policy network together with the object pose $\mathbf{x}_t$ and velocity $\mathbf{v}_t$, point cloud features $\mathbf{f}_t$, and robot proprioception $\mathbf{b}_t$ to generate the high-level action:
\begin{equation}
\mathbf{a}_t = \pi\left(\mathbf{f}_p, \mathbf{x}_t, \mathbf{v}_t, \mathbf{g}_t, \mathbf{b}_t\right).
\end{equation}

Recognizing the limitations of obtaining complete point clouds or accurate pose estimates in deployments, we design a student policy that relies only on depth and segmentation images from wrist and base-mounted cameras. A Transformer-based visual encoder captures temporal dynamics from multi-frame inputs $(\mathcal{I}_t^{\text{arm}}, \mathcal{I}_t^{\text{base}})$, and fuses them with proprioception $\mathbf{b}_t$ to generate the final action:
\begin{equation}
\mathbf{a}_t^{\text{stu}} = \pi^{\text{stu}}\left(\mathcal{I}_t^{\text{arm}}, \mathcal{I}_t^{\text{base}}, \mathbf{b}_t\right).
\end{equation}

Lastly, the student is trained to imitate the teacher by minimizing the mean squared error between their action outputs:
\begin{equation}
\mathcal{L}_{\text{KD}} = \frac{1}{T} \sum_{t=1}^{T} \left\| \mathbf{a}_t^{\text{stu}} - \mathbf{a}_t^{\text{tea}} \right\|_2^2.
\label{eq:kd_loss}
\end{equation}
This enables robust and real-time control in perception-limited scenarios, ensuring reliable dynamic grasping with minimal sensing requirements.

\subsection{High-Level Teacher Policy Network with Grasp Fusion Module}
In the simulation environment, the teacher network utilizes privileged information of the target object, including ground-truth pose $\mathbf{x}_t$ and velocity $\mathbf{v}_t$, to learn robust grasping and motion control policies. To tackle dynamic object grasping, we propose the Grasp Fusion Module (GFM) before the policy network, which has two stages: grasp memory construction and dynamic fusion.

In the \textbf{grasp memory construction stage}, under a fixed camera pose, a pretrained grasp pose prediction network (e.g., Contact-GraspNet \cite{sundermeyer2021contact}) generates $N$ grasp candidates $\{\mathbf{G}_i \in \mathbb{SE}(6)\}_{i=1}^N$ in a single forward pass. The top-$K$ candidates with highest scores are stored in a memory bank as relative transformations $\tilde{\mathbf{G}}_i$ in the object's local frame, reducing online computation.

In the \textbf{dynamic fusion stage}, the object point cloud is encoded by pretrained PointNet \cite{qi2017pointnet} into a global feature vector $\mathbf{f}_{\text{p}}$. At each time step $t$, the current object pose $\mathbf{x}_t$ and $\mathbf{f}_{\text{p}}$ form a query vector via an MLP:
\begin{equation}
\mathbf{q}_t = \mathrm{MLP}_q\left(\mathbf{f}_{\text{p}};\ \mathbf{x}_t\right).
\end{equation}

Relative transformations $\{\tilde{\mathbf{G}}_i\}$ are converted to world-frame grasps $\{\mathbf{G}_i^t\}$ using $\mathbf{x}_t$. Each $\mathbf{G}_i^t$ is flattened and mapped to key-value pairs:
\begin{equation}
\mathbf{k}_i^t = \mathrm{MLP}_k(\mathrm{vec}(\mathbf{G}_i^t)), \quad
\mathbf{v}_i^t = \mathrm{MLP}_v(\mathrm{vec}(\mathbf{G}_i^t)).
\end{equation}

An attention mechanism weights the values to produce the optimal grasp representation:
\begin{equation}
\alpha_{i,t} = \frac{\exp(\mathbf{q}_t^\top \mathbf{k}_i^t)}{\sum_{j=1}^K \exp(\mathbf{q}_t^\top \mathbf{k}_j^t)}, \quad
\mathbf{g}_t = \sum_{i=1}^K \alpha_{i,t} \mathbf{v}_i^t.
\end{equation}

Finally, the fused grasp $\mathbf{g}_t$, proprioception $\mathbf{b}_t$, point cloud feature $\mathbf{f}_{\text{p}}$, object pose $\mathbf{x}_t$, and velocity $\mathbf{v}_t$ are input to an MLP to generate high-level actions.

\subsection{High-Level Student Policy Network: Temporal Modeling with Dual-View Fusion}
We design a lightweight student policy network that takes only camera inputs. To ensure accurate control in dynamic grasping, we propose a Dual-Stream visual Modeling (DSM) structure, which fuses observations from a wrist-mounted and a base-mounted camera. This allows the policy to capture both local details and global spatial layouts.

Each view provides three consecutive frames, each containing a target mask \(\mathbf{m}_t\) (from a pretrained Track-SAM \cite{cheng2023segment}) and a depth map \(\mathbf{d}_t\). Compared to RGB, mask maps reduce the sim-to-real domain gap, while depth maps encode the 3D geometry between the camera and the target. The inputs are denoted as \(\mathcal{I}_\text{arm}\) and \(\mathcal{I}_\text{base}\).

Both input streams are passed through a shared CNN encoder \(\mathrm{CNN}_\theta\) to extract feature sequences \(\mathbf{F}^a\) and \(\mathbf{F}^b\). Meanwhile, the robot's proprioceptive state is encoded by an MLP into an embedding \(\mathbf{e}_b\), which is concatenated with each visual feature, forming fused tokens \(\tilde{\mathbf{f}}_t^a\) and \(\tilde{\mathbf{f}}_t^b\).

Temporal encodings are added, and the sequences are processed by two separate Transformer encoders (\(\mathrm{Trans}_a\), \(\mathrm{Trans}_b\)) to model time-aware attention from both views. The resulting outputs are projected via a linear layer and concatenated:
\[
\mathbf{z} = [\mathrm{Proj}(\mathrm{Trans}_a(\tilde{\mathbf{f}}^a));\ \mathrm{Proj}(\mathrm{Trans}_b(\tilde{\mathbf{f}}^b))],
\]
which is then fed into an action head to generate \(\hat{\mathbf{a}}_t\).

\subsection{Low-Level Policy For Locomotion}
The high-level policy outputs the end-effector position increment $\Delta \hat{\mathbf{p}} \in \mathbb{R}^3$, orientation increment $\Delta \hat{\mathbf{r}} \in \mathbb{R}^3$, as well as base linear velocity $v_{\text{lin}}$ and yaw rate $\omega_{\text{yaw}}$, which need to be further converted into executable low-level control signals.

The low-level controller first integrates these increments into target end-effector states $(\hat{\mathbf{p}}, \hat{\mathbf{r}})$, and combines them with base motion commands to form a low-level command vector:
\begin{equation}
\mathbf{u}_t = [\hat{\mathbf{p}}, \hat{\mathbf{r}}, v_{\text{lin}}, \omega_{\text{yaw}}] \in \mathbb{R}^8.
\end{equation}

This command vector, together with the robot’s proprioceptive state $\mathbf{b}_t$ and the terrain embedding $\mathbf{z}_t$, is fed into a multi-layer perceptron $\pi_{\text{low}}$ to output target joint angles for the legs:
\begin{equation}
\mathbf{q}^*_t = \pi_{\text{low}}(\mathbf{b}_t, \mathbf{u}_t, \mathbf{z}_t).
\end{equation}

The quadruped executes actions through PD control. For the manipulator, inverse kinematics is used to compute joint angles $\boldsymbol{\theta}^*_{\text{arm}}$ from the target end-effector state $(\hat{\mathbf{p}}_{\text{arm}}, \hat{\mathbf{r}}_{\text{arm}})$.
The final action output is:
\begin{equation}
\mathbf{a}_{low} = [\mathbf{q}^*_t, \boldsymbol{\theta}^*_{\text{arm}}].
\end{equation}

\subsection{Training Details}
To efficiently train hierarchical control policies, we use a staged training approach. First, the low-level policy is trained with PPO \cite{ppo} and then frozen. Next, a high-level teacher policy is trained via reinforcement learning using a modified reward based on static grasping. This includes a yaw-angle penalty: when the absolute yaw angle $|\psi|$ exceeds $60^\circ$, a quadratic penalty is applied; if it exceeds $70^\circ$, the episode terminates early to improve sampling efficiency. After freezing the teacher, a student policy is trained with DAgger \cite{dagger} by minimizing the MSE between student actions and teacher labels.

\begin{table*}[t]
\centering
\small
\setlength{\tabcolsep}{4pt}
\caption{Grasp Success Rate (GSR) and One-Shot Success Rate (OSSR) (\%) across four levels of increasing difficulty. Here, GSR-T and GSR-S denote the teacher and student strategies, respectively. DQ-Net consistently outperforms all baselines.}
\label{tab:combined_results}
\resizebox{\textwidth}{!}{
\begin{tabular}{
    >{\raggedright\arraybackslash}p{3.2cm}
    *{12}{S[table-format=2.1]}
}
\toprule
\multirow{2}{*}{\textbf{Method}} &
\multicolumn{3}{c}{\textbf{Level 1}} &
\multicolumn{3}{c}{\textbf{Level 2}} &
\multicolumn{3}{c}{\textbf{Level 3}} &
\multicolumn{3}{c}{\textbf{Level 4}} \\
\cmidrule(lr){2-4} \cmidrule(lr){5-7} \cmidrule(lr){8-10} \cmidrule(lr){11-13}
 & {GSR-T} & {GSR-S} & {OSSR} &
   {GSR-T} & {GSR-S} & {OSSR} &
   {GSR-T} & {GSR-S} & {OSSR} &
   {GSR-T} & {GSR-S} & {OSSR} \\
\midrule
VBC            & 0.0 & 0.0  & 0.0  & 0.0 & 0.0  & 0.0  & 0.0 & 0.0  & 0.0  & 0.0 & 0.0  & 0.0  \\
VBC-D          & 57.5 & 28.6 & 27.4 & 57.4 & 29.0 & 27.6 & 48.7 & 19.1 & 17.8 & 42.7 & 16.0 & 15.2 \\
DQ-Net w/o GFM & 65.2 & 49.8 & 44.7 & 65.2 & 49.7 & 44.8 & 60.0 & 36.0 & 31.3 & 55.0 & 33.3 & 28.9 \\
DQ-Net w/o vel & 78.5 & 47.4 & 46.9 & 78.6 & 46.9 & 46.4 & 74.0 & 38.4 & 37.6 & 70.3 & 34.8 & 34.0 \\
\rowcolor{yellow!20}
\textbf{DQ-Net}        & \textbf{80.8} & \textbf{55.8} & \textbf{53.2} & \textbf{80.8} & \textbf{55.6} & \textbf{53.2} & \textbf{77.9} & \textbf{44.8} & \textbf{41.8} & \textbf{74.3} & \textbf{41.0} & \textbf{38.5} \\
\bottomrule
\end{tabular}
}
\end{table*}

\begin{table}[t]
\centering
\caption{Grasp Success Rate (GSR) and model size comparison between student policies. Our Transformer-based method outperforms the CNN-based baseline across all difficulty levels with fewer parameters.}
\label{tab:stu_com}
\small
\setlength{\tabcolsep}{5pt}
\resizebox{\linewidth}{!}{
\begin{tabular}{
    l
    r
    r
    r
    r
    c
}
\toprule
\textbf{Method} & \textbf{L1} & \textbf{L2} & \textbf{L3} & \textbf{L4} & \textbf{Params (M)} \\
\midrule
CNN-Based & 51.61 & 51.44 & 40.13 & 36.14 & 8.43 \\
\rowcolor{yellow!20}
\textbf{Ours} & \textbf{55.82} & \textbf{55.63} & \textbf{44.87} & \textbf{41.04} & \textbf{5.37} \\
\bottomrule
\end{tabular}}
\end{table}

\section{Experiment}
\subsection{Environments}
We adopt \textbf{DQ-Bench} as the unified evaluation platform. Each method is executed for 5{,}000 steps per task level to ensure statistically stable results. All methods are trained under the Level 4 setting and are evaluated on Levels 1 through 4. To better simulate real-world conditions, the visual observations are delayed by four frames to mimic perception latency.

\subsection{Compared Methods}
\noindent\textbf{VBC \cite{liu2024vbc}.} A strong baseline for quadruped grasping static object via whole-body control.\\[0.5em]
\noindent\textbf{VBC-D.} A modified version of VBC with a reward function and angle constraints adapted for dynamic object grasping, aligned with our settings for fair comparison.\\[0.5em]
\noindent\textbf{DQ-Net w/o GFM.} A variant of DQ-Net where the GFM is removed from the teacher policy.\\[0.5em]
\noindent\textbf{DQ-Net w/o vel.} DQ-Net where object velocity is excluded from the teacher policy.

\subsection{Implementation Details}
All training and evaluation are performed on a single NVIDIA RTX 4090 GPU using the Unitree B1 quadruped robot with a Unitree Z1 robot arm. The low-level control policy is shared across all methods to ensure consistent control capability. The high-level teacher policy is trained in 6{,}000 parallel environments with a rollout length of 24, totaling 80{,}000 timesteps. The high-level student policy is trained in 200 parallel environments for the same number of timesteps. We set the number of top-ranked grasp candidates in the GFM module to $K = 30$ throughout all experiments. More details are put in the supplementary material.

\subsection{Experimental Results}
\noindent
\textbf{Grasping Performance under Varying Motion Complexity.}
We evaluated all methods across four difficulty levels to assess performance in dynamic object grasping, focusing on Grasp Success Rate (GSR). As shown in Table~\ref{tab:combined_results}, DQ-Net, under the student policy, consistently achieves the highest GSR, outperforming VBC-D by 25\% at Level 4.
Ablation variants of DQ-Net show slightly reduced performance but still surpass VBC-D across all levels. 
VBC-D, though adapted for dynamics, performs poorly due to limited motion modeling, while the static VBC nearly fails under movement. Overall, DQ-Net demonstrates strong robustness to increasing motion complexity, with GFM and velocity input each contributing to effective grasp prediction under dynamic conditions.

\textbf{Success Rate on One-Time Grasping.}
Unlike static object grasping with multiple adjustment opportunities, dynamic grasping often permits only one attempt due to object motion. Thus, the one-shot grasp success rate (OSSR) is a key metric for evaluating legged robots in dynamic scenarios. To assess each method’s decision efficiency, we compare their OSSR across five scenarios in Table~\ref{tab:combined_results}. DQ-Net consistently outperforms all baselines, demonstrating superior real-time decision-making. Even with a single-attempt constraint, it achieves strong performance across all difficulty levels.
% Unlike static object grasping, where multiple adjustments are possible, dynamic object grasping typically allows only limited or even a single attempt due to object motion. Therefore, one-shot grasp success rate (OSSR) is a critical metric for evaluating the grasping capability of legged robots in dynamic settings.
% To assess the decision efficiency and decisiveness of each method, we compare their OSSR across five scenarios, as shown in Table~\ref{tab:combined_results}. DQ-Net consistently outperforms all baselines, demonstrating superior real-time decision-making ability. Even under the constraint of single-attempt execution, DQ-Net achieves strong performance across all difficulty levels, reaching a 38.5\% success rate on the most challenging Level 4 task.

\begin{figure}[t]
    \centering
    \includegraphics[width=\linewidth]{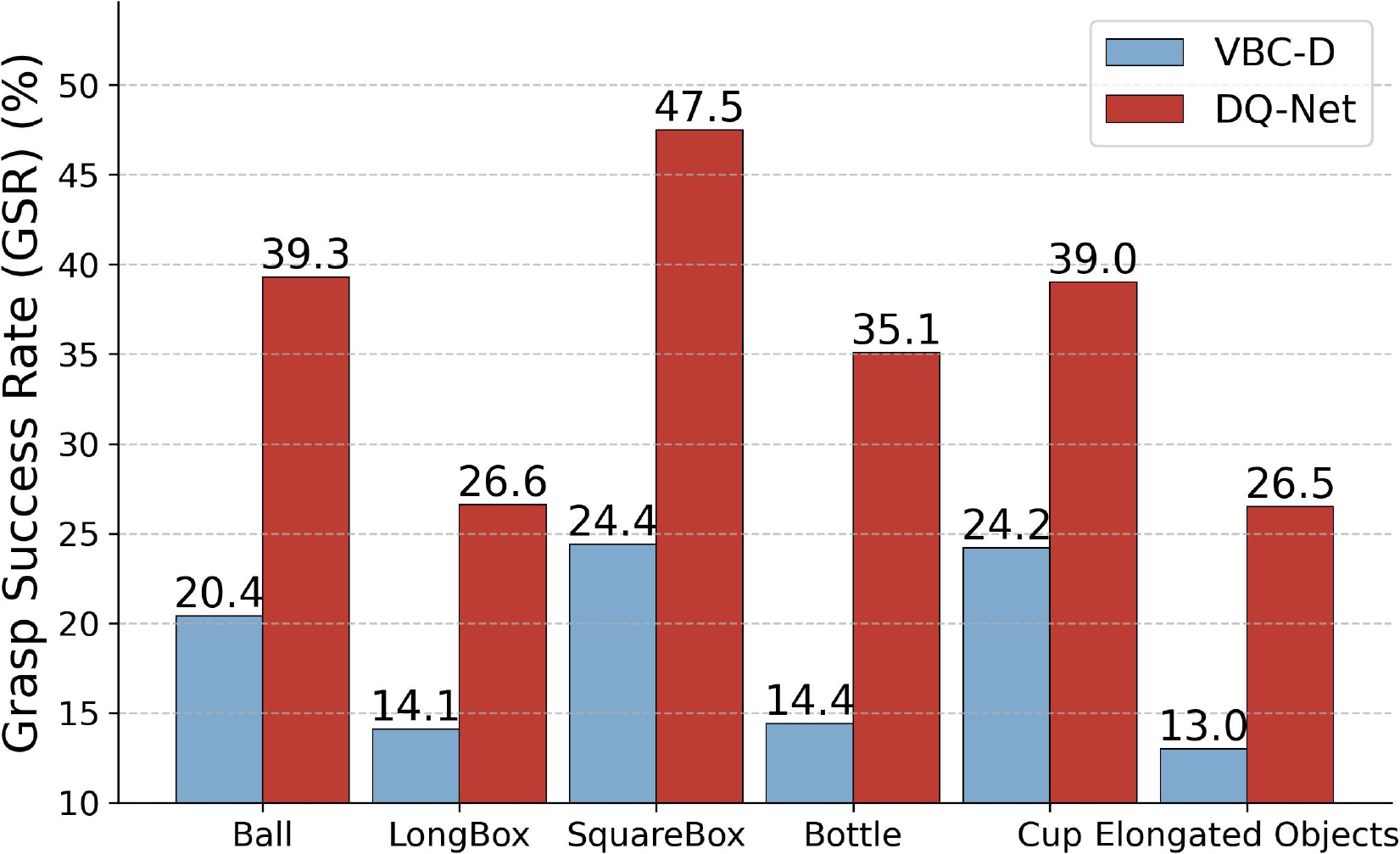}
    \caption{Grasp success rates on unseen objects under Level 4 difficulty.}
    \label{fig3:generalization}
\end{figure}

\begin{table}[t]
\centering
\captionsetup{type=table}
\caption{Average timesteps to successful completion (TSC) across four difficulty levels. Lower values indicate faster and more efficient grasping.}
\label{tab:tsc}
\small
\setlength{\tabcolsep}{3pt}
\renewcommand{\arraystretch}{1.1}
\sisetup{detect-all}
\resizebox{\linewidth}{!}{
\begin{tabular}{
    l
    S[table-format=2.2]
    S[table-format=2.2]
    S[table-format=2.2]
    S[table-format=2.2]
}
\toprule
\textbf{Method} & \textbf{Level 1} & \textbf{Level 2} & \textbf{Level 3} & \textbf{Level 4} \\
\midrule
VBC-D            & 43.71 & 43.86 & 41.20 & 41.56 \\
DQ-Net w/o GFM   & 36.20 & 35.35 & 36.91 & 35.43 \\
DQ-Net w/o vel   & 37.01 & 37.13 & 35.22 & 35.10 \\
\rowcolor{yellow!20}
\textbf{DQ-Net}  & \textbf{35.45} & \textbf{35.24} & \textbf{35.27} & \textbf{34.32} \\
\bottomrule
\end{tabular}}
\end{table}

\begin{figure*}[!t]
    \centering
    \includegraphics[width=\textwidth]{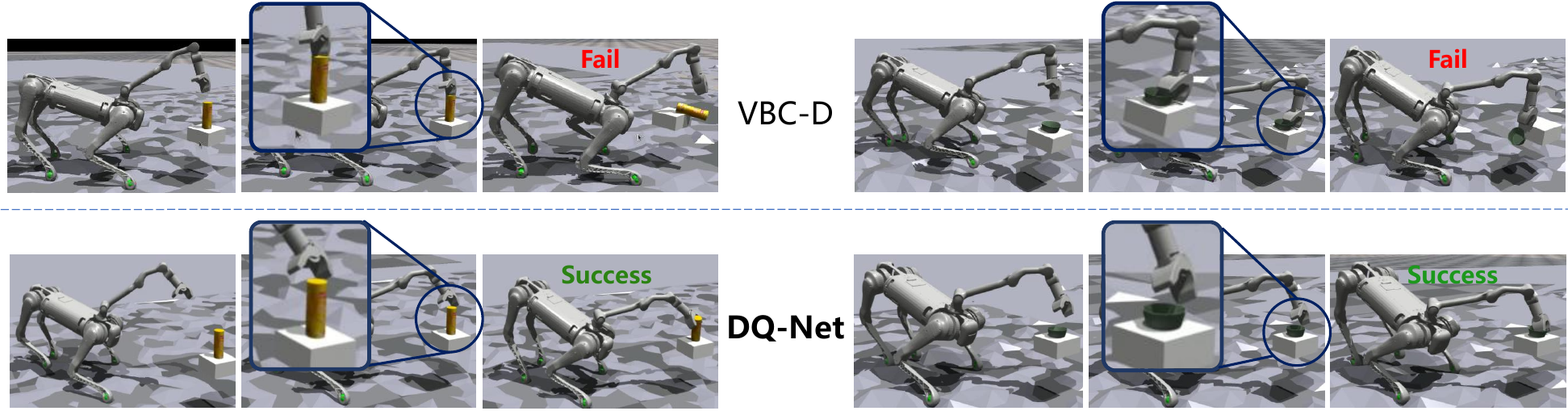}
    \caption{End-effector poses before contact in dynamic scenes. Both methods approach the object, but only DQ-Net achieves precise alignment, enabled by grasp priors from GFM. VBC-D suffers from misalignment, often leading to failure.
    }
    \label{fig:grasp_pose}
\end{figure*}

\begin{figure*}[!t]
    \centering
    \includegraphics[width=\textwidth]{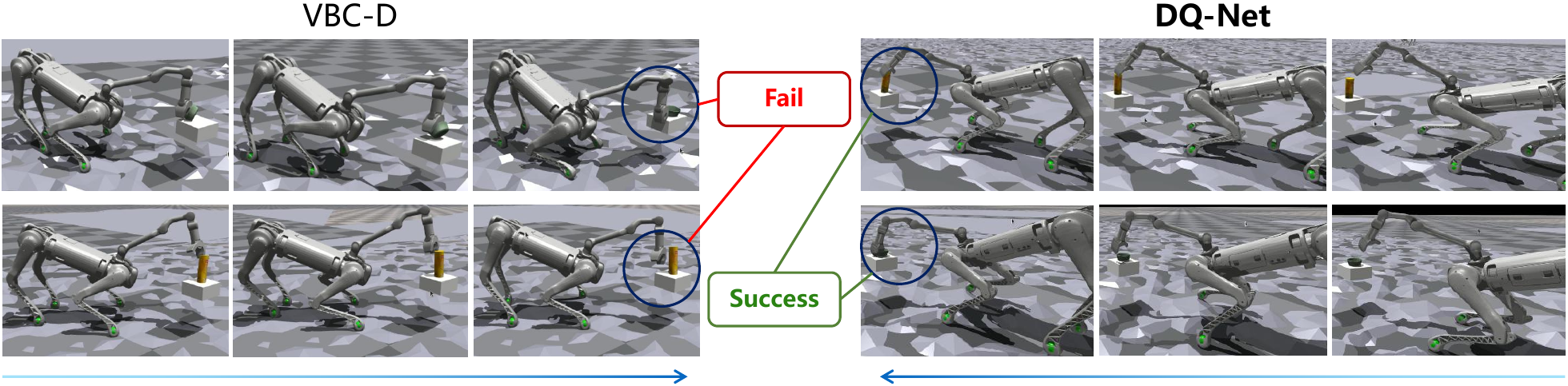}
    \caption{Comparison of motion anticipation. DQ-Net predicts future object trajectories and adjusts accordingly, while VBC-D reacts with delay. GFM and velocity cues enable DQ-Net’s predictive and adaptive grasping.
    }
    \label{fig:anticipation}
\end{figure*}

\textbf{Timesteps Comparisons.} 
To assess execution efficiency in dynamic grasping, we measure the average time steps per grasp (TSC). As shown in Table~\ref{tab:tsc}, DQ-Net and its ablated variants require significantly fewer steps than VBC-D. The three DQ-Net variants show similar TSC across all levels, indicating that both GFM and velocity input enhance object dynamics understanding and action speed. Without these cues, the policy becomes more cautious and less efficient.
% To evaluate the execution efficiency of each method in dynamic object grasping, we measure the average time steps required to complete a grasp (TSC). As shown in Table~\ref{tab:tsc}, DQ-Net and its two ablated variants require significantly fewer time steps than VBC-D, particularly in Level 1 and Level 2 tasks where VBC-D lags behind by nearly ten steps. Moreover, the three DQ-Net variants exhibit comparable TSC across all levels, suggesting that both the GFM and velocity input help the robot understand object dynamics and perform faster actions. In contrast, removing these dynamic cues leads to reduced motion understanding, causing the policy to behave more cautiously and less efficiently. 

\textbf{Generalization to Unseen Objects.}
In real-world scenarios, robots often encounter previously unseen objects, making it crucial to evaluate grasp performance on novel items to assess a policy’s generalization capability. We compare DQ-Net with the baseline VBC-D on a set of unseen YCB objects excluded from training. These objects are categorized into six types: ball, long box, square box, bottle, cup, and elongated objects (e.g., screwdriver). All evaluations are under the Level 4 setting, and grasp success rates for each category are reported. As shown in Figure~\ref{fig3:generalization}, DQ-Net consistently outperforms VBC-D, demonstrating superior generalization ability even under increased task difficulty. All detailed results on the complete set of unseen objects are provided in the appendix.

\textbf{Our approach vs. CNN-based policy for student strategy.}
We design a lightweight Transformer-based student policy that captures temporal dependencies and fine-grained visual cues via a dual-view strategy. Compared to the CNN-based student in VBC~\cite{liu2024vbc}, our model achieves consistently higher grasp success rates (Table~\ref{tab:stu_com}) with only 5.37M parameters versus 8.43M. The performance gain stems from separate modeling of camera inputs and the temporal modeling strength of the Transformer backbone.

\subsection{Qualitative Results}

To better understand the effectiveness of our method in dynamic object grasping, we visualize and compare the behavioral differences between different policies. In such tasks, both the grasping pose of the end-effector and its spatial alignment with the object's motion trajectory are crucial to grasp success. We present a qualitative comparison between VBC-D and DQ-Net in representative scenarios.

\textbf{Better Grasping Pose.}
Figure~\ref{fig:grasp_pose} illustrates the end-effector poses just before contact. While both methods bring the arm near the object, only DQ-Net aligns the gripper precisely with the target, whereas VBC-D suffers from noticeable misalignment, often leading to grasp failure. We attribute this to the GFM, which enables DQ-Net to dynamically integrate grasp priors from memory to predict a more suitable pose.

\textbf{More Predictive Pose Estimation.}
Even with a theoretically correct static grasp pose, successful grasping may still fail in dynamic scenarios if the robotic arm cannot properly adapt to the object's continuous motion.
As shown in Figure~\ref{fig:anticipation}, VBC-D exhibits delayed reactions and lacks motion anticipation, whereas DQ-Net predicts the object's future trajectory and proactively adjusts its gait and body posture. This predictive behavior fundamentally stems from DQ-Net's enhanced ability to model object motion dynamics in real time, effectively leveraging velocity cues and the GFM's memory retrieval mechanism to generate grasp poses that are optimally aligned with anticipated future object positions.

% \textbf{More Predictive Pose Estimation.}
% Even with a theoretically correct static grasp pose, successful grasping may still fail in dynamic scenarios if the robotic arm cannot properly adapt to the object's continuous motion. As clearly demonstrated in Figure~\ref{fig:anticipation}, VBC-D frequently exhibits delayed reactions and critically lacks motion anticipation capabilities when dealing with fast-moving objects. Conversely, DQ-Net demonstrates superior performance by accurately predicting the object's future trajectory and proactively adjusting both its gait pattern and body posture accordingly. This predictive behavior fundamentally stems from DQ-Net's enhanced ability to model object motion dynamics in real time, effectively leveraging velocity cues and the GFM's memory retrieval mechanism to generate grasp poses that are optimally aligned with anticipated future object positions.

\section{Conclusion}
This work presents DQ-Bench, the first benchmark for dynamic object grasping with quadruped robots, featuring realistic target motion, complex terrains, multi-level difficulty settings, and diverse object categories. Building on this benchmark, we propose DQ-Net, a unified whole-body grasping framework combining a memory-driven Grasp Fusion Module (GFM) with a lightweight student policy guided by dual-view visual observations and proprioception. DQ-Net achieves dynamic grasping by effectively fusing spatiotemporal information from both the robot and the environment.

Extensive evaluations show that DQ-Net consistently outperforms strong baselines across difficulty levels and object categories, demonstrating robust generalization and responsiveness. These results highlight DQ-Net as a promising foundation for advancing dynamic mobile manipulation.

In future work, we plan to deploy DQ-Net on real quadruped hardware, explore grasping with deformable or articulated objects, and extend the framework to multi-object and collaborative manipulation scenarios.

% \bibliographystyle{aaai2026}  
% \bibliographystyle{alpha}
% plain, unsrt, alpha, abbrv
\bibliography{aaai2026}        

\newpage
\appendix

\section{Additional Benchmark Details}
\subsection{Detailed of Used Objects}
We selected a subset of objects from the YCB dataset as training and testing data for all methods. In total, 43 object categories were used, among which 30 categories appeared during training and were designated as seen objects, while the remaining 13 categories were exclusively used for testing and categorized as unseen objects. Notably, the unseen objects differ significantly from the seen ones in terms of shape and physical attributes, rather than being merely similar variants. For instance, items such as soap and pear appear only in the unseen category and are absent from the seen set.

\begin{figure*}[!t]
    \centering
    \includegraphics[width=0.8\textwidth]{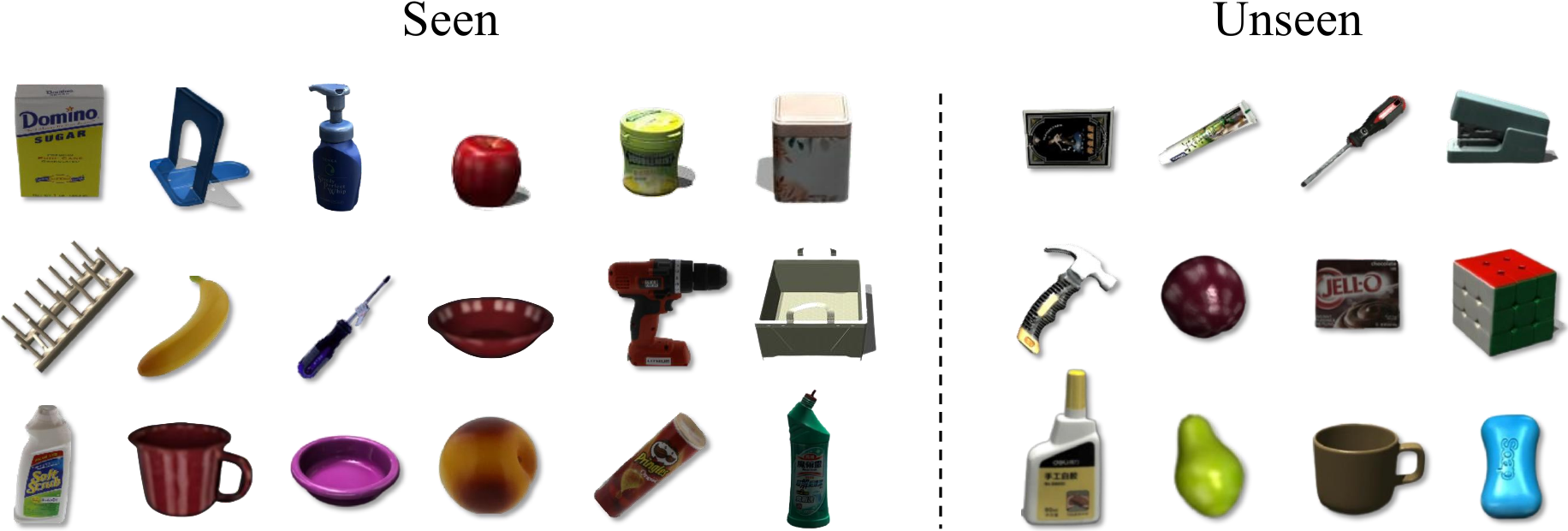}
    \caption{Examples of the seen and unseen objects.
    }
    \label{fig:objects}
    \vspace{-0.3cm}
\end{figure*}

To ensure the validity of the evaluation setup, we maintained consistency in object types across the seen and unseen groups. Specifically, both groups include short, compact objects suitable for top-down grasping and tall, slender objects better suited for frontal grasping. This setup ensures a fair and meaningful assessment of generalization performance on unseen objects under a consistent category distribution. Examples of the seen and unseen object sets are provided in Figure \ref{fig:objects}.

\subsection{Environment Setup}
DQ-Bench introduces a floating platform to induce object motion in a physically realistic manner, thereby avoiding the unrealistic aerial dynamics that may occur when applying direct forces to the object itself. Specifically, we place a floating platform in the scene and apply forces to it, enabling controlled movement in arbitrary directions and speeds. This setup lays the foundation for simulating a wide range of object motion patterns. The target object to be grasped is placed on the platform and moves as a result of the frictional and normal forces exerted by the platform, ensuring physically plausible interactions without explicitly applying forces to the object.

Furthermore, the size of the floating platform is designed to be only slightly larger than the object. We observed that when the platform area is too large, the robot tends to exploit the environment during training—for example, by gently placing the object on the platform using the manipulator and then incrementally interacting with it until a grasp is achieved. Such behavior constitutes a form of “cheating” and is generally impermissible in real-world dynamic grasping scenarios. Reducing the platform’s size ensures that sufficiently strong disturbances cause the object to fall off, thereby resetting the scene. This constraint encourages the robot to perform successful grasps in a single attempt, thus promoting more robust and generalizable grasping strategies.

\subsection{Multi-Level Motion Modes}
\begin{figure*}[!t]
    \centering
    \includegraphics[width=\textwidth]{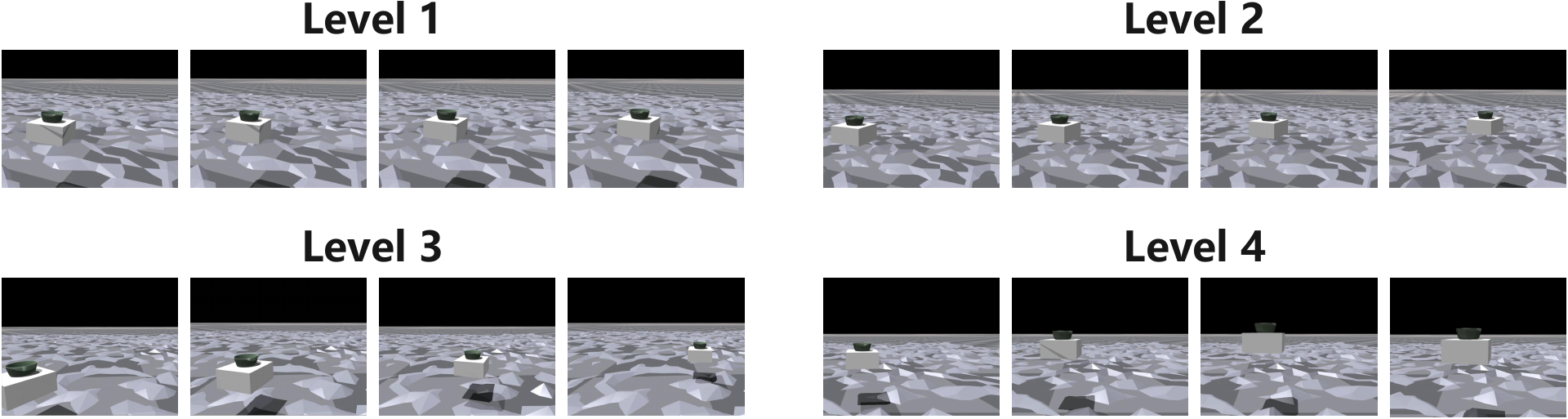}
    \caption{Motion patterns of objects across four hierarchical levels.
    }
    \label{fig:motion}
    \vspace{-0.3cm}
\end{figure*}

In DQ-Bench, we design four representative object motion modes that collectively encompass challenges related to terrain complexity, motion speed, movement patterns, and height variation. Although a wide range of motion configurations is theoretically possible, we deliberately select four modes that are highly distinguishable and well-suited for evaluating the object-grasping performance of quadruped robots in dynamic environments. All four modes are configured on rugged terrains with randomly varying elevations, thereby simulating more realistic and demanding operational conditions. Visualizations of these four motion modes are presented in Figure \ref{fig:motion}.

\section{Additional Training Details}
\subsection{Low-Level Policy Training}
To enable dexterous control of the quadruped robotic system, we adopt a general-purpose low-level control policy\cite{liu2024vbc} that facilitates whole-body motion tracking. The objective of this policy is to stably follow the velocity and end-effector pose commands issued by the high-level policy across diverse terrains and goal configurations, thereby constructing a foundational control module capable of supporting multi-task capabilities.

\textbf{Control Command Formulation.} 
The control objectives of the low-level policy encompass two aspects: velocity control of the robot base and pose control of the manipulator end-effector. Specifically, the control command is represented as
\[
\mathbf{b}_t = [\mathbf{p}_{\text{cmd}}, \mathbf{o}_{\text{cmd}}, v^{\text{lin}}_{\text{cmd}}, \omega^{\text{yaw}}_{\text{cmd}}],
\]
where $\mathbf{p}_{\text{cmd}} \in \mathbb{R}^3$ denotes the desired end-effector position, $\mathbf{o}_{\text{cmd}} \in \mathbb{R}^3$ represents the desired orientation (in Euler angles), and $v^{\text{lin}}_{\text{cmd}}$ and $\omega^{\text{yaw}}_{\text{cmd}}$ correspond to the desired linear and yaw angular velocities, all defined in the robot base frame.

\noindent
\textbf{Observation Input Structure.}
The policy receives a high-dimensional observation vector that integrates base states, manipulator joint states, leg joint states, previous actions, environmental parameters, gait phase information, and current control commands, expressed as:
\[
\mathbf{o}_t = [\mathbf{s}^{\text{base}}_t, \mathbf{s}^{\text{arm}}_t, \mathbf{s}^{\text{leg}}_t, \mathbf{a}_{t-1}, \mathbf{z}_t, \mathbf{t}_t, \mathbf{b}_t],
\]
where the environmental parameters $\mathbf{z}_t$ are introduced to enhance policy generalization across environments with varying physical properties. During training, we randomize physical parameters such as mass, friction coefficients, and inertial properties in simulation to generate diverse scenarios. These randomized parameters are embedded into $\mathbf{z}_t$ to improve the policy's domain robustness. This design is inspired by the "external state encoder" mechanism in works such as Rapid Motor Adaptation\cite{kumar2021rma}. The observation includes the current orientation, joint positions and velocities, contact states, and a phase vector used to maintain periodic gaits (e.g., trot gait).

\noindent
\textbf{Action Output Modality.} 
The low-level policy directly outputs target joint angles for all 12 leg joints, which are subsequently converted into torque commands via Proportional-Derivative (PD) controllers. For manipulator control, inverse kinematics is employed to compute the target joint angles based on the desired end-effector pose. Specifically, the incremental joint configuration $\Delta \boldsymbol{\theta}$ is computed using the pseudoinverse of the Jacobian matrix:
\[
\Delta \boldsymbol{\theta} = J^\top(JJ^\top)^{-1} \mathbf{e},
\]
where $J$ is the Jacobian matrix and $\mathbf{e}$ denotes the pose error between the current and desired end-effector states.

\noindent
\textbf{Target Sampling Strategy.} 
To enhance the generalizability and workspace coverage of the policy, target poses for the end-effector are randomly generated during training. To ensure consistency and spatial coverage, target positions are sampled in a local coordinate frame with fixed height, roll, and pitch. The target position is represented in spherical coordinates $(l, \phi, \psi)$ and interpolated to form a smooth trajectory:
\[
\mathbf{p}_{\text{cmd},t} = \frac{t}{T} \mathbf{p}_\text{end} + \left(1 - \frac{t}{T}\right)\mathbf{p},
\]
where $\mathbf{p}_\text{end}$ is the newly sampled target point and $T$ denotes the total trajectory duration. If the sampled target results in a collision or exceeds the workspace of the manipulator, a new sample is immediately drawn.

\begin{table}[t]
\centering
\caption{Definition of Symbols}
\resizebox{0.9\linewidth}{!}{
\begin{tabular}{ll}
\toprule
\textbf{Name} & \textbf{Symbol} \\
\midrule
Leg joint positions & $\mathbf{q}$ \\
Leg joint velocities & $\dot{\mathbf{q}}$ \\
Leg joint accelerations & $\ddot{\mathbf{q}}$ \\
Target leg joint positions & $\mathbf{q}^*$ \\
Leg joint torques & $\boldsymbol{\tau}$ \\
Base linear velocity & $\mathbf{v}_b$ \\
Base angular velocity & $\boldsymbol{\omega}_b$ \\
Base linear velocity command & $\mathbf{v}_x^*$ \\
Base angular velocity command & $\mathbf{v}_{\text{yaw}}^*$ \\
Number of collisions & $n_c$ \\
Feet contact force & $\mathbf{f}^{\text{foot}}$ \\
Feet velocity in Z axis & $\mathbf{v}_z^{\text{foot}}$ \\
Feet air time & $t_{\text{air}}$ \\
Timing offsets & $\boldsymbol{\theta}_{\text{cmd}}$ \\
Stepping frequency & $f_{\text{cmd}}$ \\
\bottomrule
\label{tab:symbols}
\end{tabular}
}
\end{table}

\begin{table*}[t]
\centering
\caption{Reward Function Details for the Low-Level Policy}
\resizebox{0.85\textwidth}{!}{
\begin{tabular}{lll}
\toprule
\textbf{Name} & \textbf{Definition} & \textbf{Weight} \\
\midrule
Linear velocity tracking & $\phi(\mathbf{v}_{b,xy}^* - \mathbf{v}_{b,xy})$ & 1 \\
Yaw velocity tracking & $\phi(\mathbf{v}_{\text{yaw}}^* - \omega_b)$ & 0.5 \\
Angular velocity penalty & $- \|\boldsymbol{\omega}_{b,xy}\|^2$ & 0.05 \\
Joint torques & $- \|\boldsymbol{\tau}\|^2$ & 0.00002 \\
Action rate & $- \|\mathbf{q}^*\|^2$ & 0.25 \\
Collisions & $- n_{\text{collision}}$ & 0.001 \\
Feet air time & $\sum_{i=0}^{4} (t_{\text{air},i} - 0.5)$ & 2 \\
Default Joint Position Error & $\exp(-0.05\|\mathbf{q} - \mathbf{q}_{\text{default}}\|)$ & 1 \\
Linear Velocity z & $\|\mathbf{v}_{b,z}\|^2$ & $-1.5$ \\
Base Height & $\|h_b - h_{b,\text{target}}\|$ & $-5.0$ \\
Swing Phase Tracking (Force) & $\sum_{\text{foot}} \left(1 - C_{\text{foot}}^{\text{cmd}}(\boldsymbol{\theta}_{\text{cmd}}, t)\right) \left(1 - \exp\left(-\|(\mathbf{f}_{\text{foot}})^2 / \sigma_{cf}\| \right)\right)$ & $-0.2$ \\
Stance Phase Tracking (Velocity) & $\sum_{\text{foot}} (C_{\text{foot}}^{\text{cmd}}(\boldsymbol{\theta}_{\text{cmd}}, t)) \left(1 - \exp\left(-\|(\mathbf{v}_z^{\text{foot}})^2 / \sigma_{cv} \|\right)\right)$ & $-0.2$ \\
\bottomrule
\label{tab:low_reward}
\end{tabular}
}
\end{table*}

\noindent
\textbf{Reward Design and Optimization.} 
The policy is trained via reinforcement learning with a composite reward function incorporating multiple components, like VBC\cite{liu2024vbc}, including command tracking, energy penalty, smooth locomotion, and gait phase synchronization, as detailed in Table~\ref{tab:symbols} and \ref{tab:low_reward}. The desired velocity tracking component employs a Gaussian kernel to measure the deviation between actual and target velocities, complemented by additional terms such as contact stability, action smoothness, and base height constraints. These collectively promote a balance between stability and motion flexibility.

\subsection{High-Level Policy Training}
\noindent
\textbf{GFM Details.}
In the Teacher Policy, we design an attention fusion module to perform weighted selection of candidate grasp poses based on the query features. This module takes as input the candidate grasp poses $\texttt{grasps} \in \mathbb{R}^{B \times N \times 6}$, the point cloud features $\texttt{obj\_feat} \in \mathbb{R}^{B \times 128}$, and the object pose $\texttt{pose} \in \mathbb{R}^{B \times 6}$. The point cloud features $\texttt{obj\_feat}$ and the object pose $\texttt{pose}$ are concatenated and projected via a linear layer $\texttt{Linear}(134 \rightarrow 64)$ to obtain the query vector $\texttt{Q} \in \mathbb{R}^{B \times 1 \times 64}$. Each candidate grasp is projected through another linear layer $\texttt{Linear}(6 \rightarrow 64)$ to produce the key and value vectors, denoted as $\texttt{K}, \texttt{V} \in \mathbb{R}^{B \times N \times 64}$. The attention scores between $\texttt{Q}$ and $\texttt{K}$ are computed and normalized via a Softmax operation to obtain the attention weights. These weights are then used to compute a weighted sum of $\texttt{V}$, resulting in the fused grasp representation $\texttt{Weighted} \in \mathbb{R}^{B \times 1 \times 64}$. Finally, the fused feature is projected through a linear layer $\texttt{Linear}(64 \rightarrow 6)$ to generate the final grasp pose in the shape of $\mathbb{R}^{B \times 6}$, thereby enabling optimal grasp selection conditioned on task context information.

\noindent
\textbf{Reward Functions.}
In this project, the high-level policy training follows the reward structure proposed for static grasping tasks in VBC\cite{liu2024vbc}. The overall reward is composed of two main categories: Task-oriented Rewards and Assistant Rewards. The task-oriented component adopts a staged design, including an approach reward \( r_{\text{approach}} \), a lifting reward \( r_{\text{lift}} \) (corresponding to \( r_{\text{progress}} \)), and a task completion reward \( r_{\text{completion}} \). Only one stage-specific reward is activated at a time based on the robot's current state, ensuring clear decoupling between different task phases.

For the assistant rewards, the design objectives include enhancing policy smoothness, accelerating training convergence, and improving sim-to-real transferability. We retain several original components, such as the joint velocity penalty \( r_{\text{acc}} \), action smoothness reward \( r_{\text{action}} \), end-effector orientation alignment \( r_{\text{ee\_orn}} \), base orientation alignment \( r_{\text{base\_orn}} \), and base approach reward \( r_{\text{base\_approach}} \). In addition, we introduce several new reward terms tailored to the task requirements and adjust the weights of certain existing components:

\begin{itemize}
    \item \textbf{Base Height Constraint Reward \( r_{\text{base\_h}} \)}: Encourages the robot to maintain a reasonable base height during the grasping task. Defined as \( r_{\text{base\_h}} = \exp(-|H_c - H_t|) \), where \( H_c \) and \( H_t \) are the current and target heights, respectively;
    \item \textbf{Yaw Angle Penalty \( r_{\text{yaw\_pen}} \)}: Applies a nonlinear penalty when the yaw deviation \( |\Psi_c - \Psi_0| \) exceeds \( \pi/3 \), formulated as \( r_{\text{yaw\_pen}} = -\tanh(|\Psi_c - \Psi_0|) \), which helps keep the robot aligned with the task direction;
    \item Additionally, to further constrain the yaw angle, we incorporate a quadratic penalty mechanism inspired by \cite{liu2024vbc}:
    \begin{equation}
        r_{\text{yaw}} = -\mathbb{I}(|\psi_c - \psi_0| > \pi/3) \cdot \tanh(|\psi_c - \psi_0|),
    \end{equation}
    where \( \psi \) is the current yaw angle. This penalty is triggered when \( |\psi| > 60^\circ \), and episodes are terminated early if \( |\psi| > 70^\circ \) to improve sample efficiency.
\end{itemize}

The full list of reward terms and their associated weights is provided in Table~\ref{tab:reward_weights}.

\begin{table*}[t]
\centering
\caption{High-Level Reward components and their associated weights.}
\renewcommand{\arraystretch}{1.4}
\begin{tabular}{l|l|l}
\hline
\textbf{Category} & \textbf{Definition} & \textbf{Weight} \\
\hline

\multirow{3}{*}{Task-oriented Rewards} 
& \( r_{\text{approach}} \) & 0.5 \\
\cline{2-3}
& \( r_{\text{lift}} \) & 0.8 \\
\cline{2-3}
& \( r_{\text{completion}} \) & 3.5 \\
\hline

\multirow{8}{*}{Assistant Rewards} 
& \( r_{\text{acc}} = 1 - \exp\left(-\left\| \dot{\mathbf{q}}_{t-1} - \dot{\mathbf{q}}_t \right\|\right) \) & -0.001 \\
\cline{2-3}
& \( r_{\text{cmd}} = -\left\| v_x^* \right\| + 0.25 \cdot \exp\left(-\left\| v_x^* \right\|\right) \) & 0.05 \\
\cline{2-3}
& \( r_{\text{action}} = 1 - \exp\left(-\left\| \mathbf{a}_{t-1} - \mathbf{a}_t \right\|\right) \) & -0.001 \\
\cline{2-3}
& \( r_{\text{ee\_orn}} = \cos\left( \angle(\mathbf{d}_{\text{obj}}, \mathbf{d}_{\text{ee}}) \right) \) & 0.01 \\
\cline{2-3}
& \( r_{\text{base\_orn}} = \cos\left( \angle(\mathbf{d}_{\text{obj}}, \mathbf{d}_{\text{base}}) \right) \) & 0.25 \\
\cline{2-3}
& \( r_{\text{base\_approach}} = 1 + \tanh\left( -10 \cdot \left\| x_{\text{obj}} - x_{\text{base}} - 0.6 \right\| \right) \) & 0.01 \\
\cline{2-3} 
& \( r_{\text{base\_h}} = \exp\left(-\left| H_c - H_t \right|\right) \) & 0.5 \\
\cline{2-3} 
& \( r_{\text{yaw}} = -\tanh(|\Psi_c - \Psi_0|)\text{, if } |\Psi_c - \Psi_0| > \pi/3 \) & -0.4 \\
\hline
\end{tabular}
\label{tab:reward_weights}
\end{table*}

\noindent
\textbf{Student policy network details.}
DQ-Net adopts a dual-stream Transformer-based architecture that integrates visual observations and robot state information for high-level decision-making. The overall network comprises a shared convolutional encoder, guided Transformer modules, and an action regression head. At each timestep, the observation is represented by paired mask and depth images captured from both the manipulator and base perspectives. Each input includes three historical frames, resulting in a tensor of shape $[B, 12, 54, 96]$, where 12 corresponds to two modalities, two viewpoints, and three temporal frames. The input images are first processed by a shared CNN encoder to extract low-dimensional per-frame features. This CNN consists of two convolutional layers with kernel sizes 5 and 3, respectively, followed by ELU activation, pooling, and fully connected layers, ultimately encoding each frame into a 64-dimensional feature vector. The visual features from the manipulator and base are then arranged into separate feature sequences of length three and fed into two independent guided Transformer modules. Each Transformer contains two encoder layers with two attention heads per layer. A linearly projected robot state token is prepended to the input sequence to guide the Transformer, and positional encoding is employed to enhance temporal modeling. The output visual tokens are averaged to obtain a fused representation. Finally, the features from the manipulator and base streams are projected to a unified dimension, concatenated and passed through a three-layer fully connected regression head with ELU activation to predict the target actions.

\section{Additional Experimental Details}
\subsection{Transformer-based vs. CNN-based Student Policy}
\begin{figure*}[!t]
    \centering
    \includegraphics[width=0.85\textwidth]{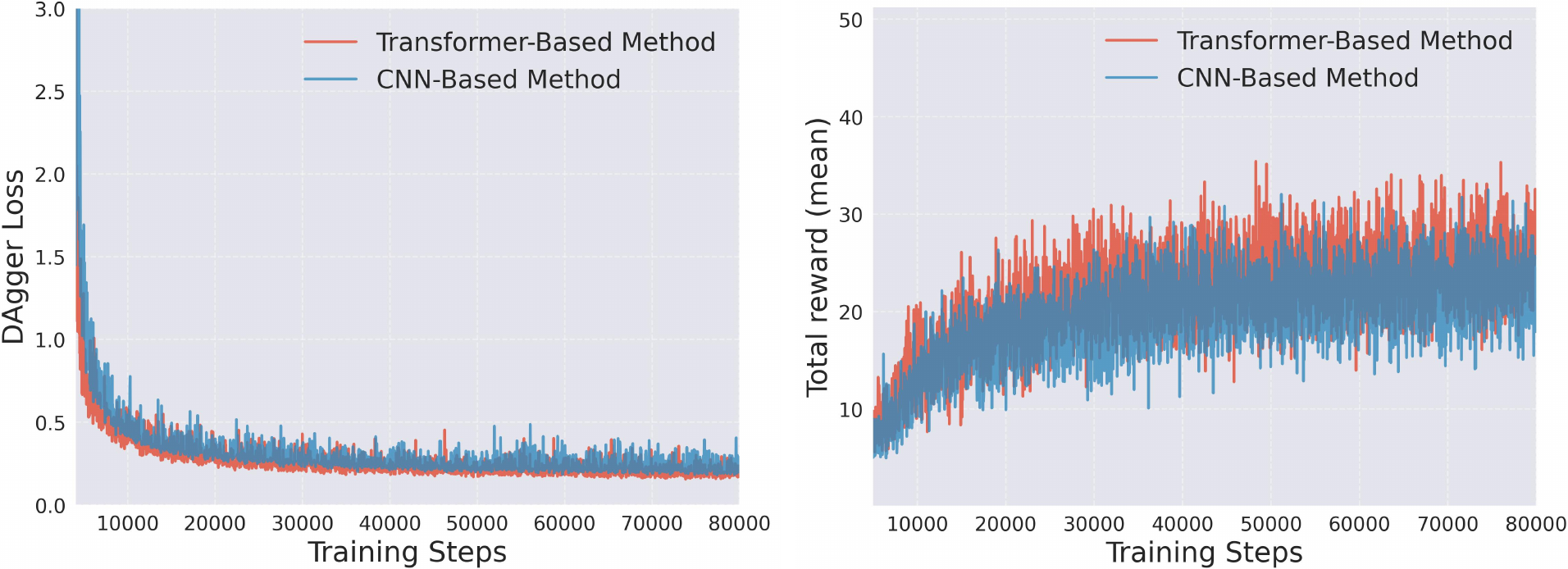}
    \caption{Comparison of Dagger Loss and Total Reward (mean) between Transformer-Based and CNN-Based Policy.
    }
    \label{fig:stu_com}
    \vspace{-0.3cm}
\end{figure*}

We propose a lightweight student policy based on a Transformer architecture capable of modeling temporal dependencies. By incorporating a dual-view modeling strategy, our approach effectively captures fine-grained visual details from the manipulator-mounted camera. To validate the effectiveness of our student design, we compare it against the CNN-based student architecture used in VBC \cite{liu2024vbc}. Specifically, both student networks are distilled from an identical teacher policy with comparable performance, and trained for 80,000 steps. Figure \ref{fig:stu_com} presents the loss curves and mean total reward\cite{zeng2024video2reward} trajectories during the distillation process for both architectures. Our model not only converges faster with lower training loss, but also achieves consistently higher reward values. These results further highlight the advantages of temporal modeling and fine-detail visual feature extraction, enabling our student policy to better approximate the teacher’s behavior. Ultimately, our approach demonstrates superior performance in dynamic object manipulation tasks, confirming its suitability for quadrupedal robots operating in complex, real-world environments.

\subsection{Comprehensive Comparison of Grasping Success Rates on Unseen Objects}
\begin{figure*}[!t]
    \centering
    \includegraphics[width=0.85\textwidth]{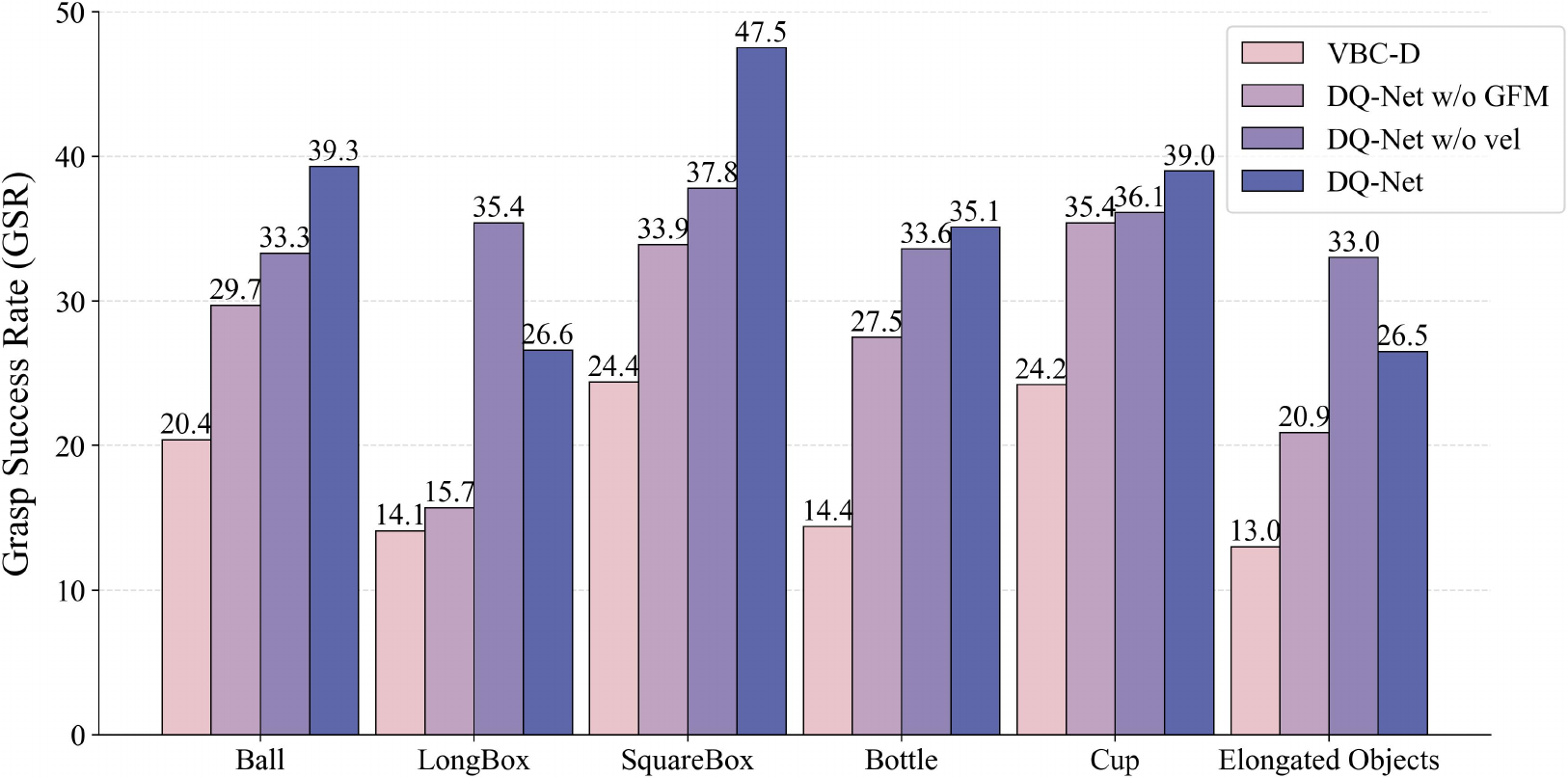}
    \caption{Grasp Prediction Visualization for Objects in IsaacGYM.
    }
    \label{fig:unseen_all}
    \vspace{-0.3cm}
\end{figure*}

% 我们对所有方法在未见物体上的抓取成功率进行了比较，实验结果如 Figure \ref{unseen_all} 所示。值得注意的是，剔除了物体速度输入但保留抓取融合模块的 DQ-Net 变体（DQ-Net w/o vel），在某些未见物体类别上的表现甚至优于完整模型。这一现象表明，尽管显式的运动信息有助于在训练分布中准确建模目标轨迹，但也可能使策略过拟合于特定的运动模式，从而在面对分布外的新物体时削弱其泛化能力。完整 DQ-Net 借助 6D 速度信息实现更强的运动预测能力，但在遇到几何形状或运动规律显著不同的物体时，这种依赖可能反而成为性能瓶颈。相比之下，DQ-Net w/o vel 更依赖几何特征与由抓取融合模块（GFM）提供的抓取先验，通过维护一个基于对象几何构建的抓取记忆库，并在执行阶段融合当前感知信息，从而生成稳健的抓取姿态。在缺乏速度信息的情况下，模型倾向于通过空间对齐和几何一致性来完成抓取，进而展现出更强的泛化能力和鲁棒性。这一发现表明，在泛化至关重要的场景中，适当降低对动态信息的依赖，反而能够提升策略的稳定性与跨类别的迁移能力，也进一步凸显了在设计动态抓取策略时，对运动感知与几何推理之间平衡的重视。
We compare the grasp success rate of all methods on unseen objects, as illustrated in Figure \ref{fig:unseen_all}. Interestingly, the variant of DQ-Net that removes explicit object velocity input while retaining the Grasp Fusion Module (DQ-Net w/o vel) occasionally outperforms the full model on certain unseen object categories. This counterintuitive result suggests that, although explicit motion information can aid in trajectory prediction within familiar settings, it may inadvertently lead to policy overfitting to the motion patterns seen during training, thereby impairing generalization to novel object dynamics. The full DQ-Net leverages 6-DoF velocity inputs to enable anticipatory and trajectory-consistent grasp predictions; however, such reliance can become a liability when confronted with objects exhibiting unfamiliar shapes or motion characteristics. In contrast, DQ-Net w/o vel places greater emphasis on geometric reasoning and grasp priors retrieved via the Grasp Fusion Module (GFM), which maintains a memory bank of high-quality grasp candidates based solely on object geometry. During execution, these priors are dynamically fused with current perceptual observations to produce robust grasp poses. In the absence of velocity cues, the policy learns to generalize through spatial alignment and geometric consistency, resulting in improved robustness and adaptability to previously unseen objects. These findings highlight that, in scenarios where generalization is critical, reducing dependence on temporally variable or privileged information can enhance the stability and transferability of grasping policies. Moreover, this underscores the importance of carefully balancing motion awareness and geometric understanding when designing strategies for dynamic object manipulation in open-world environments.

\subsection{Comparative Analysis of Teacher Policy Performance via Grasp Success Rate}
\begin{table*}[t]
\centering
\caption{Comparison of Grasp Success Rate (GSR, mean ± std) for teacher policies across four difficulty levels in DQ-Bench.}
\label{tab:teacher_gsr}
\small
\setlength{\tabcolsep}{4pt} % 增加列间距
\resizebox{0.7\textwidth}{!}{ % 撑满宽度
\begin{tabular}{l 
    S[table-format=2.1]@{\,±\,}S[table-format=1.2] 
    S[table-format=2.1]@{\,±\,}S[table-format=1.2]
    S[table-format=2.1]@{\,±\,}S[table-format=1.2]
    S[table-format=2.1]@{\,±\,}S[table-format=1.2]
}
\toprule 
\textbf{Method} & \multicolumn{2}{c}{\textbf{Level 1}} & \multicolumn{2}{c}{\textbf{Level 2}} & \multicolumn{2}{c}{\textbf{Level 3}} & \multicolumn{2}{c}{\textbf{Level 4}} \\
\midrule
VBC-D           & 57.5 & 2.7 & 57.4 & 2.8 & 48.7 & 2.6 & 42.7 & 2.9 \\
DQ-Net w/o GFM  & 65.2 & 4.1 & 65.2 & 4.1 & 60.0 & 4.8 & 55.0 & 4.7 \\
DQ-Net w/o vel  & 78.5 & 3.8 & 78.6 & 3.7 & 74.0 & 3.3 & 70.3 & 3.5 \\
\textbf{DQ-Net} & \textbf{80.8} & \textbf{1.7} & \textbf{80.8} & \textbf{1.8} & \textbf{77.9} & \textbf{1.4} & \textbf{74.3} & \textbf{1.2} \\
\bottomrule
\end{tabular}
}
\end{table*}
%为深入分析教师网络设计中各关键组件的作用，我们在 DQ-Bench 提供的四级动态抓取任务中，对比评估了不同策略下教师网络的抓取成功率（Grasp Success Rate, GSR）。
% 我们共评估四种教师策略：经典的基线方法 VBC-D、DQ-Net 移除 GFM 模块的变体（DQ-Net w/o GFM）、DQ-Net 去除目标速度信息的变体（DQ-Net w/o vel）以及完整的 DQ-Net。所有策略均在 DQ-Bench 的 Level 4 环境中进行训练，并分别在 Level 1–4 四个难度等级下测试抓取性能，以保证比较的公平性与一致性。实验结果如表~\ref{tab:tsc} 所示。
% 从结果可见，完整的 DQ-Net 在所有难度等级下均取得了最优表现，Level 4 下 GSR 达到 $74.3\% \pm 1.2$，显著优于其余策略。这表明 GFM 与目标速度信息共同为高质量抓取姿态的预测提供了关键支持。相比之下，移除目标速度（DQ-Net w/o vel）后，策略在高动态等级中表现出明显退化，说明目标运动状态对抓取策略的实时调整至关重要。而在无 GFM 的情形下，策略缺乏先验抓取知识的动态整合能力，导致抓取质量与一致性大幅度下降，特别是在任务复杂度提升时更为明显，相较于DQ-Net下降了\%19.3。
% 此外，VBC-D 尽管引入了对动态任务的适配修改，但其性能仍远低于 DQ-Net 系列。我们推测其原因在于：该方法缺乏面向目标运动的时序建模能力与显式抓取候选整合机制，导致策略在目标快速运动或轨迹不确定的情况下难以有效生成高质量抓取动作。。
To thoroughly investigate the role of critical components within the teacher network, we conduct a comparative evaluation of Grasp Success Rate (GSR) across four levels of dynamic grasping tasks provided by DQ-Bench. This analysis focuses on four teacher policy variants: the adapted baseline VBC-D, a version of DQ-Net with the Grasp Fusion Module (GFM) removed (DQ-Net w/o GFM), a version without access to object velocity (DQ-Net w/o vel), and the full DQ-Net implementation.

All teacher policies are trained under the most challenging Level 4 setting to ensure robustness, and then evaluated on Levels 1 through 4 to ensure a fair and consistent comparison. The experimental results are summarized in Table~\ref{tab:teacher_gsr}.

As shown in the results, the full DQ-Net achieves the highest GSR across all difficulty levels, reaching $74.3\% \pm 1.2$ on Level 4, significantly outperforming all ablated variants. This indicates that both the Grasp Fusion Module and object velocity information play a crucial role in enabling accurate and stable grasp pose estimation under dynamic conditions. In contrast, removing velocity input (DQ-Net w/o vel) leads to noticeable degradation in performance under high-motion scenarios, underscoring the importance of motion-aware policy adjustment. Furthermore, without the GFM (DQ-Net w/o GFM), the policy loses its capacity to dynamically incorporate prior grasp knowledge, resulting in a substantial performance drop, especially as task complexity increases—showing a 19.3\% decline in GSR compared to the full DQ-Net.

Despite the dynamic adaptation in VBC-D, its performance remains significantly lower than that of all DQ-Net variants. We hypothesize this is due to the lack of temporal modeling for object motion and the absence of an explicit grasp proposal integration mechanism. As a result, VBC-D struggles to generate reliable grasp actions when faced with rapid or unpredictable object trajectories.

\section{Limitations and Future Work}
Despite the impressive performance of DQ-Net in dynamic object grasping tasks, several limitations remain that warrant further investigation. First, the current framework relies on knowledge distillation, where the student network learns the teacher's strategy. However, in highly dynamic and unpredictable environments, object motion patterns are diverse and complex, resulting in a high-dimensional action space. The student network, constrained by limited perceptual inputs (e.g., target masks and depth maps), struggles to replicate the teacher's decisions that are based on privileged information such as precise object pose and velocity. This leads to a noticeable performance gap, particularly in scenarios involving fast-moving or irregular trajectories, where the student’s grasp success rate and responsiveness are suboptimal. Second, the current Grasp Fusion Module (GFM) depends on a static, pre-defined set of grasp candidates. Although an attention mechanism is employed to select high-quality grasps dynamically, the fixed candidate set limits adaptability to novel objects or extreme motion patterns. Additionally, GFM's limited generalization capability can result in suboptimal decisions in out-of-distribution scenarios. To address these challenges, we propose a hybrid training framework that integrates reinforcement learning (RL) with knowledge distillation. The student network will refine its policy through trial-and-error, guided by task-specific reward functions tailored for dynamic settings—such as grasp success rate, motion smoothness, and energy efficiency—thus enhancing its adaptability and narrowing the performance gap with the teacher. Furthermore, to improve the adaptability and generalization of the GFM, we plan to introduce an online learning mechanism that enables real-time updates of the grasp candidate set. By leveraging self-supervised learning during robot-environment interaction, the system can continuously extract novel grasp patterns, thereby dynamically expanding and optimizing the candidate pool. This approach is expected to significantly enhance the system’s robustness and effectiveness in handling unseen scenarios.

\section{Extended Related Work}
\textbf{Dynamic Object Grasping.}

Dynamic object grasping has attracted increasing attention due to its broad applications in industrial automation and human–robot collaboration. Traditional approaches typically adopt a perception–planning decoupled paradigm, where the object motion is first estimated and predicted, followed by the generation of time-parameterized grasping trajectories. For instance, Menon et al.~\cite{menon2014motion} formulate dynamic grasping as a motion planning problem under dynamic constraints and leverage motion primitives to rapidly generate interception trajectories. While effective in structured environments, such methods generally assume a static manipulator, limiting the operational workspace and rendering them inadequate for high-speed or large-scale target motion.

To overcome the limitations of modular frameworks, recent works have explored end-to-end learning approaches. GAP-RL~\cite{xie2024gap} encodes grasp poses using 3D Gaussian representations and jointly optimizes 6-DoF grasp detection and policy learning, achieving strong robustness across diverse dynamic trajectories. GraspARL~\cite{wu2022grasparl} adopts adversarial training, where an adversary perturbs object trajectories to train robust grasping policies. Although these methods achieve promising results in one-shot dynamic grasping, they still rely on static bases and lack scalability to large-scale or distributed tasks.

In addition, perception-driven methods emphasizing temporal consistency have gained attention. Liu et al.~\cite{liu2023target} propose a target-referenced grasping strategy based on graph neural networks, enabling semantic-consistent transfer of grasp poses across video frames for continuous object tracking. However, their approach assumes fixed viewpoints and close-range scenarios, highlighting the limitations of non-mobile platforms in dynamic settings.

To address spatial constraints, Zhang et al.~\cite{zhang2023dynamic} introduce a quadruped-robot-based dynamic grasping framework. They propose a visual servoing strategy based on a spherical projection model, enabling monocular RGB cameras to estimate and track unknown object velocities for dynamic grasping. Nevertheless, the system lacks full-body coordination control, limiting its adaptability to varying object heights. Moreover, experiments are conducted only on pre-defined motion trajectories, without validation under random target motion.

In summary, although recent advances in dynamic grasping have achieved progress in integrating perception, prediction, and policy learning, most existing methods are constrained by static platforms or insufficient whole-body control. These limitations result in restricted spatial coverage and poor adaptability to complex, unstructured dynamic grasping tasks.
% \subsection{Legged Manipulator}

\section{Additional Details of Grasping Network}
\begin{figure}[t]
    \centering
    \includegraphics[width=0.8\columnwidth]{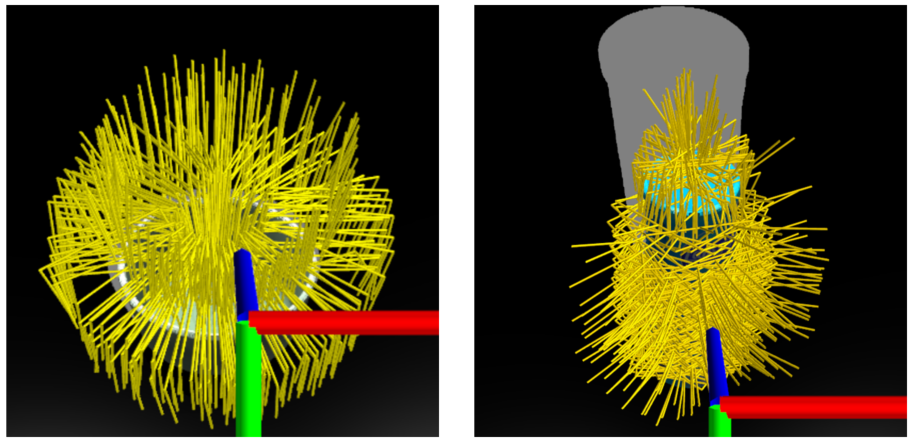}
    \caption{Grasp Prediction Visualization for Objects in IsaacGYM.
    }
    \label{fig:contact_graspnet}
    \vspace{-0.3cm}
\end{figure}

Contact-GraspNet\cite{sundermeyer2021contact} is an end-to-end framework that directly predicts 6-DoF grasp poses from depth images in complex environments, particularly suited for multi-view grasping of unknown objects. By anchoring the grasp representation on visible contact points in the point cloud, it projects the full 6-DoF pose onto the observable surface.
% effectively reducing the high-dimensional regression problem to the prediction of grasp orientation and width. 
% This significantly simplifies the learning process. 
In our project, Contact-GraspNet is integrated into the Isaac Gym environment to generate multi-view and multi-pose grasp predictions for target objects (as shown in Figure \ref{fig:contact_graspnet}). The predicted results exhibit high diversity in grasp poses, providing a rich and stable set of candidate grasps for the teacher network, thereby facilitating subsequent pose fusion by the GFM.

% \bibliographystyle{unsrt}
% \bibliography{aaai2026}

\end{document}